\documentclass{article}
\usepackage[font={footnotesize,sf},labelfont={footnotesize,sf,bf},skip=2.5pt]{caption}
\usepackage[accepted]{icml2018}
\usepackage{microtype}
\usepackage{graphicx}
\usepackage{subfigure}
\usepackage{booktabs}
\usepackage{amsfonts} 
\usepackage{amsmath,amssymb}
\usepackage{relsize}
\usepackage{cleveref}
\newcommand{\eg}{e.g.\ }

\newcommand{\flabel}[1]{\label{fig:#1}}
\newcommand{\seclabel}[1]{\label{sec:#1}}
\newcommand{\tlabel}[1]{\label{tab:#1}}
\newcommand{\elabel}[1]{\label{eq:#1}}
\newcommand{\alabel}[1]{\label{alg:#1}}

\newcommand{\fref}[1]{\Cref{fig:#1}}
\newcommand{\sref}[1]{\Cref{sec:#1}}
\newcommand{\tref}[1]{\Cref{tab:#1}}

\newcommand{\figdir}{figures/}
\newcommand{\capt}[2]{\caption[#1.]{\textbf{#1.}#2}}

\newcommand{\fig}[5]
{
\begin{figure}
\begin{center}
\includegraphics[width=#3\columnwidth]{\figdir/#1}
\end{center}
\capt{#4}{#5}
\flabel{#2}
\end{figure}
}

\newcommand{\figstar}[5]
{
\begin{figure*}
\begin{center}
\includegraphics[width=#3\textwidth]{\figdir/#1}
\end{center}
\capt{#4}{#5}
\flabel{#2}
\end{figure*}
}

\newcommand{\twofigstarH}[5]
{
\begin{figure}[H]
\begin{center}
\begin{minipage}{\columnwidth}
\begin{center}
\includegraphics[width=\textwidth]{\figdir/#1} 
\end{center}
\end{minipage}

\vspace{0.1cm}

\begin{minipage}{\columnwidth}
\begin{center}
\includegraphics[width=\textwidth]{\figdir/#2} 
\end{center}
\end{minipage}
\end{center}
\capt{#4}{#5}
\flabel{#3}
\end{figure}
}
\newcommand{\twofigstar}[5]
{
\begin{figure}
\begin{center}
\begin{minipage}{\columnwidth}
\begin{center}
\includegraphics[width=\textwidth]{\figdir/#1} 
\end{center}
\end{minipage}

\vspace{0.1cm}

\begin{minipage}{\columnwidth}
\begin{center}
\includegraphics[width=\textwidth]{\figdir/#2} 
\end{center}
\end{minipage}
\end{center}
\capt{#4}{#5}
\flabel{#3}
\end{figure}
}

\newcommand{\expect}[3]{\mathop{\mathlarger{\mathbb{E}}}_{#1 \sim #2}\left[#3\right]}
\newcommand{\expectshrink}[3]{\mathop{\mathlarger{\mathbb{E}}}_{#1 \sim #2}\!\!\!\!\!\!\!\left[#3\right]}

\begin{document}

\twocolumn[
\icmltitle{Associative Compression Networks for Representation Learning}

\begin{icmlauthorlist}
\icmlauthor{Alex Graves}{deepmind}
\icmlauthor{Jacob Menick}{deepmind}
\icmlauthor{A\"aron van den Oord}{deepmind}
\end{icmlauthorlist}

\icmlaffiliation{deepmind}{DeepMind, London, UK}

\icmlkeywords{Deep Learning, Generative Models, Variational Autoencoders, Compression}

\vskip 0.3in
]

\printAffiliationsAndNotice{}

\begin{abstract}
This paper introduces Associative Compression Networks (ACNs), a new framework for variational autoencoding with neural networks.
The system differs from existing variational autoencoders (VAEs) in that the prior distribution used to model each code is conditioned on a similar code from the dataset.
In compression terms this equates to sequentially transmitting the dataset using an ordering determined by proximity in latent space.
Since the prior need only account for local, rather than global variations in the latent space, the coding cost is greatly reduced, leading to rich, informative codes.
Crucially, the codes remain informative when powerful, autoregressive decoders are used, which we argue is fundamentally difficult with normal VAEs.
Experimental results on MNIST, CIFAR-10, ImageNet and CelebA show that ACNs discover high-level latent features such as object class, writing style, pose and facial expression, which can be used to cluster and classify the data, as well as to generate diverse and convincing samples.
We conclude that ACNs are a promising new direction for representation learning: one that steps away from IID modelling, and towards learning a structured description of the dataset as a whole.
\end{abstract}

\section{Introduction}
\label{Introduction}
Unsupervised learning---the discovery of structure in data without extrinsic reward or supervision signals---is likely to be critical to the development of artificial intelligence, as it enables algorithms to exploit the vast amounts of data for which such signals are partially or completely lacking.
In particular, it is hoped that unsupervised algorithms will be able to learn compact, transferable representations that will benefit the full spectrum of cognitive tasks, from low-level pattern recognition to high-level reasoning and planning.

Variational Autoencoders (VAEs) \cite{kingma2013auto,rezende2014stochastic} are a class of generative model in which an encoder network extracts a stochastic code from the data, and a decoder network then uses this code to reconstruct the data.
From a representation learning perspective, the hope is that the code will provide a high-level description or abstraction of the data, which will guide the decoder as it models the low-level details.
However it has been widely observed (\eg \citealt{chen2016variational,van2017neural}) that sufficiently powerful decoders---especially autoregressive models such as pixelCNN~\cite{oord2016conditional}---will simply ignore the latent codes and learn an unconditional model of the data.
Authors have proposed various modifications to correct this shortcoming, such as reweighting the coding cost~\cite{higgins2017beta} or removing it entirely from the loss function~\cite{van2017neural}, weakening the decoder by e.g. limiting its range of context~\cite{chen2016variational,bowman2015generating} or adding auxiliary objectives that reward more informative codes, for example by maximising the mutual information between the prior distribution and generated samples \cite{zhao2017infovae}---a tactic that has been fruitfully applied to Generative Adversarial Networks~\cite{chen2016infogan}.
These approaches have had considerable success at discovering useful and interesting latent representations.
However they add parameters to the system that must be tuned by hand (e.g. weightings for various terms in the loss function, domain specific limitations on the decoder etc.) and in most cases yield worse log-likelihoods than purely autoregressive models.

To understand why VAEs do not typically improve on the modelling performance of autoregressive networks, it is helpful to analyse the system from a minimum description length perspective~\cite{chen2016variational}.
In that context a VAE embodies a two-part compression algorithm in which the code for each datum in the training set is first transmitted to a receiver equipped with a prior distribution over codes, followed by the residual bits required to correct the predictions of the decoder (to which the receiver also has access).
The expected transmission cost of the code (including the `bits back' received by the posterior;~\citealt{hinton1993keeping}) is equal to the Kullback-Leibler divergence between the prior and the posterior distribution yielded by the encoder, while the residual cost is the negative log-likelihood of the data under the predictive distribution of the decoder.
The sum of the two, added up over the training set, is the compression cost optimised by the VAE loss function\footnote{We ignore for now the description length of the prior and the decoder weights, noting that the former is likely to be negligible and the latter could be minimised with \eg variational inference~\cite{hinton1993keeping,graves2011practical}}.

The underlying assumption of VAEs is that transmitting a piece of high-level information, for example that a particular MNIST image represents the digit 3, will be outweighed by the increased compression of the data by the decoder.
But this assumption breaks down if the decoder is able to learn a distribution that closely matches the density of the data.
In this case, if one-tenth of the training images are 3's, finding out that a particular image is a 3 will only save the decoder around $\log_2 10$ bits.
Furthermore, since an accurate prior will give a ten percent probability to 3's, it will cost exactly the same amount for the encoder to transmit that information via the prior.
In practice, since the code is stochastic and the decoder is typically deterministic, it is often more efficient to ignore the code entirely.

If we follow the above reasoning to its logical conclusion we come to a paradox that appears to undermine not only VAEs, but any effort to use high-level concepts to compress low-level data: the benefit of associating a particular concept with a particular piece of data will always be outweighed by the coding cost.
The resolution to the paradox is that high-level concepts become efficient when a single concept can be collectively associated with many low-level data, rather than pointed to by each datum individually.
This suggests a paradigm where latent codes are used to organise the training set as a whole, rather than annotate individual training examples. 
To return to the MNIST example, if we first sort the images according to digit class, then transmit all the zeros followed by all the ones and so on, the cost of transmitting the places where the digit class changes will be negligible compared to the cumulative savings over all the images of each class.
Conversely, consider an encyclopaedia that has been carefully structured into topics, articles, paragraphs and so on, providing high-level context that is known to lead to improved compression.
Now imagine that the encyclopaedia is transmitted in a randomly ordered sequence of 100 character chunks, attached to each of which is a code specifying the exact place in the structure from which it was drawn (topic X, article Y, paragraph Z etc.).
It should be clear that this would be a very inefficient compression algorithm; so inefficient, in fact, that it would not be worth transmitting the structure at all.

Encyclopaedias are already ordered, and the most efficient way to compress them may well be to simply preserve the ordering and use an autoregressive model to predict one token at a time.
But in general we do not know how the data should be ordered for efficient compression.
It would be possible to find such an ordering by minimising a similarity metric defined directly on the data, such as Euclidean distance in pixel space or edit distance for text; however such metrics tend to be limited to superficial similarities (in the case of pixel distance we provide evidence of this in our experiments).
We therefore turn to the similarity, or \emph{association}~\cite{bahdanau2014neural,graves2014neural}, among latent representations to guide the ordering.
Transmitting associated codes consecutively will only be efficient if we have a prior that captures the local statistics of the area they inhabit, and not the global statistics of the entire dataset: if a series of pictures of sheep has just been sent, the prior should expect another sheep to come next.
We achieve this by using a neural network to condition the prior on a code chosen from the $K$ nearest neighbours in latent space to the code being transmitted.
Previous work has considered fitting mixture models as VAE priors~\cite{nalisnick2016approximate,tomczak2017vae}, and one could think of our procedure as fitting a conditional prior to a uniform mixture over the $K$ posterior codes closest to whichever code we are about to transmit.
As $K \rightarrow N$, the size of the training set, we recover the familiar setting of fitting an unconditional prior.
Among supervised methods, perhaps the closest point of reference is Matching Networks~\cite{vinyals2016matching} in which a nearest neighbours search over embeddings is leveraged for one-shot learning.

Conditioning on neighbouring codes does not obviously lead to a compression procedure.
However we can define the following sequential compression algorithm if we insist that the neighbour for each code in the training set is unique:
\begin{itemize}
\item
Alice and Bob share the weights of the encoder, decoder and prior networks\footnote{In normal VAEs the encoder does not need to be shared}.
\item
Alice chooses an ordering for the training set, then transmits one element at a time by sending first a sample from the encoding distribution, then the residual bits required for lossless decoding.
\item
After decoding each data sample, Bob re-encodes the data using his copy of the encoder network, then passes the statistics of the encoding distribution into the prior network as input. The resulting prior distribution is used to transmit the next code sample drawn by Alice at a cost equal to the KL between their distributions\footnote{The prior for the first example may be assumed to be shared at negligible cost for a large dataset}.
\end{itemize}
The optimal ordering Alice should choose is the one that minimises the sum of the KLs at each transmission step.
Finding this ordering is a hard optimisation problem in general, but our empirical results suggest that the KL cost of the optimal ordering is well approximated by $K$ nearest neighbour sampling, given a suitable value of $K$.

It should be clear that ACNs are not IID in the usual sense: they optimise the cost of transmitting the entire dataset, in an order of their choosing, as opposed to the expected cost of transmitting a single data-point.
One consequence is that the ACN loss function is not directly comparable to that of VAEs or other generative models.
Indeed, since the expected cost of transmitting a uniformly random ordering of a size $N$ dataset is $\log_2{N!}$ bits, it be could argued that an ACN has $\mathcal{O}(\log_2{N})$ `free bits' per data-point to spend on codes relative to an IID model.
However, we contend that it is exactly the information contained in the ordering, or more generally in the relational structure of dataset elements, that defines the high-level regularities we wish our representation to capture.
For example, if half the voices in a speech database are male and half are female, compression should be improved by grouping according to gender, motivating the inclusion of gender in the latent codes; likewise representing speaker characteristics should make it possible to co-compress similar voices, and if there were enough examples of the same or similar phrases, it should become advantageous to encode linguistic information as well.

As the relationship between a particular datum and the rest of the dataset is not accessible to the decoder in an ACN, there is no need to weaken the decoder; indeed we recommend using the most powerful decoder possible to ensure that the latent codes are not cluttered by low-level information.
Similarly, there is no need to modify the loss function or add extra terms to encourage the use of latent codes.
Rather, the use of latent information is a natural consequence of the separation between high-level relations \emph{among} data, and low-level dependencies \emph{within} data.
As our experiments demonstrate, this leads to compressed representations that capture many salient features likely to be useful for downstream tasks.

\section{Background: Variational Auto-Encoders}
Variational Autoencoders (VAEs)~\cite{kingma2013auto,rezende2014stochastic} are a family of generative models consisting of two neural networks
---an encoder and a decoder---trained in tandem.
The encoder receives observable data $x$ as input and emits as output a data-conditional distribution $q(z|x)$ over latent vectors $z$.
A sample $z \sim q$ is drawn from this distribution and used by the decoder to determine a code-conditional reconstruction distribution $r(x|z)$ over the original data~\footnote{We use $r(x|z)$ instead of the usual notation $p(x|z)$ to avoid confusion with the ACN prior}.
The VAE loss function is defined as the expected negative log-likelihood of $x$ under $r$ (often referred to as the reconstruction cost) plus the KL divergence from some prior distribution $p(z)$ to $q(z|x)$ (referred to as the KL or coding cost):
\begin{equation*}
L^{VAE}(x) = KL(q(z|x)||p(z)) - \expect{z}{q}{\log r(x|z)}   
\end{equation*}
Although VAEs with discrete latent variables have been explored~\cite{mnih2016variational}, most are continuous to allow for stochastic backpropagation using the reparameterisation trick~\cite{kingma2013auto}.
The prior $p$ may be a simple distribution such as a unit variance, zero mean Gaussian, or something more complex such as an autoregressive distribution whose parameters are adapted during training~\cite{chen2016variational,gulrajani2016pixelvae}.
In all cases however, the prior is constant for all $x$.

\section{Associative Compression Networks}\seclabel{acn}
Associative compression networks (ACNs) are similar to VAEs, except the prior for each $x$ is now conditioned on the distribution $q(z|\hat{x})$ used to encode some neighbouring datum $\hat{x}$.
We used a unit variance, diagonal Gaussian for all encoding distributions, meaning that $q(z|x)$ is entirely described by its mean vector $\expect{z}{q(z|x)}{z}$, which we refer to as the \emph{code} $c$ for $x$.
Given $c$, we randomly pick $\hat{c}$, the code for $\hat{x}$, from $\textrm{KNN}(x)$, the set of $K$ nearest Euclidean neighbours to $c$ among all the codes for the training data.
We then pass $\hat{c}$ to the prior network to obtain the conditional prior distribution $p(z|\hat{c})$ and hence determine the KL cost.
Adding this KL cost to the usual VAE reconstruction cost yields the ACN loss function:
\begin{align*}\elabel{acn_loss}
L^{ACN}(x) = \!\!\!\!\expectshrink{\hat{c}}{\textrm{KNN}(x)}{KL(q(z|x)||p(z|\hat{c})} - \expect{z}{q}{\log r(x|z)}
\end{align*}
As with normal VAEs, the prior distribution may be chosen from a more or less flexible family.
However, as each local prior is already conditioned on a nearby code, the marginal prior across latent space will be highly flexible even if the local priors are simple.
For our experiments we chose an independent mixture prior for each dimension of latent space, to encourage multimodal but independent (and hence, hopefully, disentangled) representations.

As discussed in the introduction, conditioning on neighbouring codes is equivalent to a sequential compression algorithm, as long as every neighbour is unique to a particular code.
This can be ensured by a simple modification to the above procedure: restrict $\textrm{KNN}(x)$ at each step to contain only codes that have not yet been used as neighbours during the current pass through the dataset. 
With $K=1$ this is equivalent to a greedy nearest neighbour heuristic for the Euclidean travelling salesman problem of finding the shortest tour through the codes.
The route found by this heuristic may be substantially longer than the optimal tour, which in any case may not correspond to the ordering that minimises the KL cost, as this depends on the KLs between the priors and the codes, and not directly on the distance between the codes.
Nonetheless it provides an upper bound on the optimal KL cost, and hence on the compression of the dataset (note that the reconstruction cost does not depend on the ordering, as the decoder is conditioned only on the current code).
We provide results in \sref{experiments} to calibrate the accuracy of $L^{ACN}$ against the KL cost yielded by an actual tour.

To optimise $L^{ACN}$ we create an associative dataset $\mathbf{C}$ that holds a separate code vector $c(x)$ for each $x$ in the training set $\mathbf{X}$ and run the following algorithm:
\begin{algorithm}[H]
   \caption{Associative Compression Network Training}
   \alabel{acn}
\begin{algorithmic}
   \STATE {\bfseries Initialise $\mathbf{C}$}: $c(x) \sim \mathcal{N}(0, 1)\ \forall x \in \mathbf{X}$
   \REPEAT
   \STATE Sample $x$ uniformly from $\mathbf{X}$
   \STATE Run encoder network, get $q(z|x)$
   \STATE Update $\mathbf{C}$ with new code: $c(x) \leftarrow \expect{z}{q(z|x)}{z}$
   \STATE $\textrm{KNN}(x) \leftarrow K$ nearest Euc. neighbours to $c(x)$ in $\mathbf{C}$\label{knn}
   \STATE Pick $\hat{c}$ randomly from $\textrm{KNN}(x)$
   \STATE Run prior network, get $p(z|\hat{c})$
   \STATE $z \sim q(z|x)$
   \STATE Run decoder network, compute $\log r(x|z)$
   \STATE $L^{ACN}(x) = KL(q(z|x)||p(z|\hat{c})) - \log r(x|z)$
   \STATE Compute gradients, update network weights
   \UNTIL{convergence}
\end{algorithmic}
\end{algorithm}
In general $x$, $c(x)$ and $\hat{c}$ will be batches computed in parallel.
As the codes in $\mathbf{C}$ are only updated when the corresponding data is sampled, the codes used for the $K$ nearest neighbour (KNN) search will in general be somewhat stale.
To check that this wasn't a significant problem, we ran tests in which a parallel worker continually updated the codes using the current weight for the encoder network.
For our experiments increasing the code-update frequency made no discernible difference to learning; however code staleness could become more damaging for larger datasets.
Likewise the computational cost of performing the KNN search was low compared to that of activating the networks for our experiments, but could become prohibitive for large datasets.

\subsection{Unconditional Prior}\seclabel{unconditional}
Unlike normal VAEs, ACNs by default lack an unconditional prior, which makes it difficult to compare them to existing generative models.
However we can easily fit an unconditional prior $p(z)$ to samples drawn from the codes in $\mathbf{C}$ after training is complete.

\subsection{Sampling}\seclabel{sampling}
There are several different ways to sample from ACNs, of which we consider three.
Firstly, by drawing a latent vector $z$ from the unconditional prior $p(z)$ defined above and sampling from the decoder distribution $r(x|z)$, we can generate \emph{unconditional} samples that reflect ACNs global data distribution.
Secondly, by choosing a batch of real images, encoding them and decoding conditioned on the resulting code, we can generate stochastic \emph{reconstructions} of the images, revealing which features of the original are represented in the latents and transmitted to the decoder.
Note that in order to reduce sampling noise we use the mean codes $c$ as latents for the reconstructions, rather than samples from $\mathcal{N}(c, 1)$; we assume at this point that the decoder is autoregressive.
Lastly, the use of conditional priors opens up an alternative sampling protocol, where sequences of linked samples are generated from real data by iteratively encoding the data, sampling from the prior conditioned on the code, generating new data, then encoding again.
We refer to these sequences as `daydreams', as they remind us of the chains of associative imagining followed by the human mind at rest.
The daydream sampling process is illustrated in \fref{daydream_sampling}.

\fig{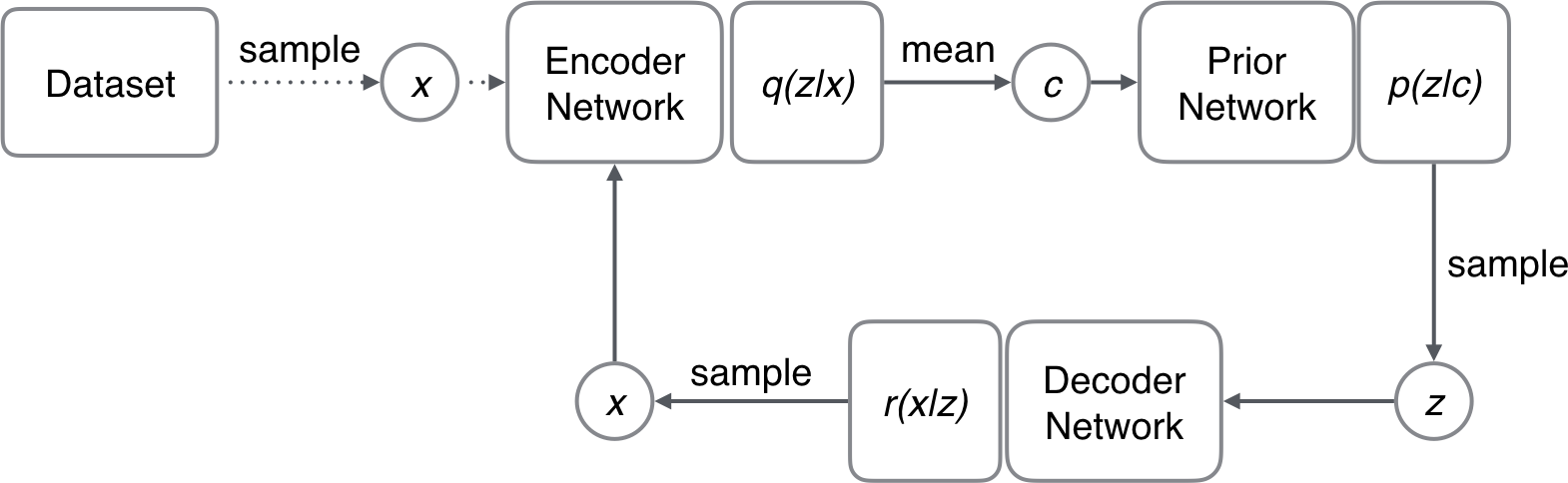}{daydream_sampling}{1}{Flow diagram for daydream sampling}{}

\subsection{Test Set Evaluation}\seclabel{test_set}
Since the true KL cost depends on the order in which the data is transmitted, there are some subtleties in comparing the test set performance of ACN with other models.
For one thing, as discussed in the introduction, most other models are order-agnostic, and hence arguably due a refund for the cost of specifying an arbitrary ordering (in the case of MNIST this would amount to 8.21 nats per test set image).
We can resolve this by calculating both an upper bound on the \emph{ordered} compression yielded by ACN, and the \emph{unordered} compression which can be computed using the KL between the unconditional prior $p(z)$ discussed in \sref{unconditional} and the test set encodings (recall that the reconstruction cost is unaffected by the ordering).
As well as providing a fair comparison with previous results, the unconditional KL gives an idea of the total amount of information encoded for each data point, relative to the dataset as a whole.
Another issue is that if an ordering is used, it is debatable whether the training and test set should be compressed together, with a single tour through all the data, or whether the test set should be treated as a separate tour, with the prior network conditioned on test set codes only.
We chose the latter for simplicity, but note that doing so may unrealistically inflate the KL costs; for example if the test set is dramatically smaller than the training set, and the average distance between codes is correspondingly larger, the density of the prior distributions may be strongly miscalibrated.

\section{Experimental Results}
\seclabel{experiments}
 
We present experimental results on four image datasets: binarized MNIST~\cite{salakhutdinov2008quantitative}, CIFAR-10~\cite{krizhevsky2009learning}, ImageNet~\cite{deng2009imagenet} and CelebA~\cite{liu2015deep}.
Buoyed by our belief that the latent codes will not be ignored no matter how well the decoder can model the data, we used a Gated PixelCNN decoder~\cite{oord2016conditional} to parameterise $p(x|z)$ for all experiments.
The ACN encoder was a convolutional network fashioned after a VGG-style classifier~\cite{simonyan2014very}, and the encoding distribution $q(z|x)$ was a unit variance Gaussian with mean specified by the output of the encoder network. 
The prior network was an MLP with three hidden layers each containing 512 $\tanh$ units, and skip connections from the input to all hidden layers and all hiddens to the output layer.
The ACN prior distribution $p(z|c)$ was parameterised using the outputs of the prior network as follows:
\begin{equation*}
p(z|c) = \prod_{d=1}^D \sum_{m=1}^M \pi^d_m \mathcal{N}(z^d|\mu^d_m, \sigma^d_m),
\end{equation*}
where $D$ was the dimensionality of $z$, $z^d$ is the $d^{th}$ element of $z$, there are $M$ mixture components for each dimension, and all parameters $\pi^d_m, \mu^d_m, \sigma^d_m$ are emitted by the prior network, with the softmax function used to normalise $\pi^d_m$ and the softplus function used to ensure $\sigma^d_m > 0$.
We used $M=8$ for MNIST and $M=16$ elsewhere; the results did not seem very sensitive to this.
Polyak averaging~\cite{polyak1992acceleration} was applied for all experiments with a decay parameter of 0.9999; all samples and test set costs were calculated using averaged weights.
For the unconditional prior $p(z)$ we always fit a Gaussian mixture model using Expectation-Maximization, with the number of components optimised on the validation set.

For all experiments, the optimiser was rmsprop~\cite{tieleman2012lecture} with learning rate $10^{-5}$ and momentum $0.9$.
The encoding distribution $q(z|x)$ was always a unit variance Gaussian with mean specified by the output of the encoder network. 
The dimensionality of $z$ was $16$ for binarized MNIST and $128$ otherwise.
Unless stated otherwise, $K=5$ was used for the KNN lookups during ACN training.

\subsection{Binarized MNIST}
For the binarized MNIST experiments the ACN encoder had five convolutional layers, and the decoder consisted of 10 gated residual blocks, each using 64 filters of size 5x5. 
The decoder output was a single Bernoulli distribution for each pixel, and a batch size of 64 was used for training.

\begin{table}
\caption{Binarized MNIST test set compression results}
\tlabel{mnist_compression_results}
\vskip 0.15in
\begin{center}
\begin{footnotesize}
\begin{sc}
\begin{tabular}{lr}
\toprule
Model & Nats / image \\
\midrule
Gated Pixel CNN (ours)   & 81.6 \\
Pixel CNN~\cite{oord2016pixel}    & 81.3 \\
Discrete VAE~\cite{rolfe2016discrete} & 81.0 \\
DRAW~\cite{gregor2015draw} & $\leq 81.0$ \\
G. PixelVAE  ~\cite{gulrajani2016pixelvae} & 79.5 \\
Pixel RNN~\cite{oord2016pixel}    & 79.2 \\
VLAE~\cite{chen2016variational} & 79.0 \\
GLN~\cite{veness2017online} & 79.0 \\
\textbf{MatNet~\cite{bachman2016architecture}}& $\mathbf{\leq 78.5}$ \\
\midrule
ACN (unordered) & 80.9 \\
\bottomrule
\end{tabular}
\end{sc}
\end{footnotesize}
\end{center}
\vskip -0.1in
\end{table}

\begin{table}
\caption{Binarized MNIST test set ACN costs}
\tlabel{mnist_acn_costs}
\vskip 0.15in
\begin{center}
\begin{footnotesize}
\begin{sc}
\begin{tabular}{lr}
\toprule
Cost & Nats / image \\
\midrule
KL ($K=1$)         & 2.6 \\
KL ($K=5$)         & 3.5 \\
KL (Greedy Tour)    & 3.6 \\
KL ($K=10$)        & 4.1 \\
KL (Unconditional) & 10.6 \\
\midrule
Reconstruction   & 70.3 \\
\midrule
ACN (ordered) & $\leq$ 73.9 \\
\bottomrule
\end{tabular}
\end{sc}
\end{footnotesize}
\end{center}
\vskip -0.1in
\end{table}
The results in \tref{mnist_compression_results} show that unordered ACN gives similar compression to the decoder alone (Gated Pixel CNN), supporting the thesis that conventional VAE loss is not significantly reduced by latent codes when using an autoregressive decoder.
\tref{mnist_acn_costs} shows that the upper bound on the ordered ACN cost (sum of greedy tour KL and reconstruction) is 7 nats per image lower than the unordered ACN cost.
Given that the cost of specifying an ordering for the test set is 8.21 nats per image, this suggests that the model is using most of the `free bits' to encode latent information.
The KL cost yielded by the `greedy tour' heuristic described in \sref{acn} is close to that given by KNN sampling on the test set codes with $K=5$ (note that we are varying $K$ when computing the test set KL only; the network was trained with $K=5$).
Since this is a loose upper bound on the optimal KL for an ordered tour, and since the $K=1$ result is a lower bound (no tour can do better than always hopping to the nearest neighbour) we speculate that the true KL is somewhere between $K=1$ and $K=5$.

\fig{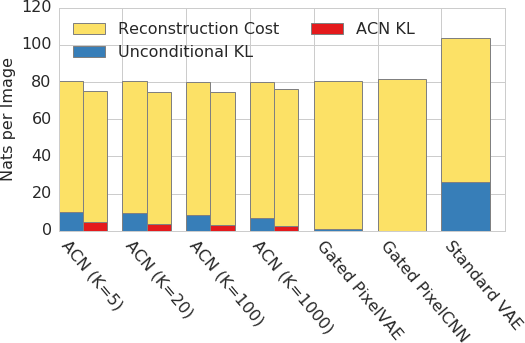}{bin_mnist_compression}{0.9}{MNIST compression costs}{ Unordered compression cost is height of blue and yellow bar, ordered compression cost (ACN models only) is height of red and yellow bar.}

As discussed in the introduction, if the value $K$ for the KNN lookups approaches the size of the training set, ACN should reduce to a VAE with a learned prior.
To test this, we trained ACNs with $K=5$ to $1000$, and measured the change in compression costs.
We also implemented a standard feedforward VAE and a VAE with the same encoder and decoder as ACN, but with an unconditional Gaussian mixture prior whose parameters were trained in place of the prior network.
We refer to the latter as \emph{Gated PixelVAE} due to similarity with previous work~\cite{gulrajani2016pixelvae}; but note that they used a fixed prior and a somewhat different encoder architectures.
\fref{bin_mnist_compression} shows that the unordered compression cost per test set image is much the same for ACN regardless of $K$, and very similar to that of both Gated PixelVAE and Gated PixelCNN (again underlining the marginal impact of latent codes on VAE loss).
However the distribution of the costs changes, with higher reconstruction cost and lower KL cost for higher $K$.
As predicted, Gated PixelVAE performs similarly to ACN with very high $K$.
The VAE performs considerably worse due to the non-autoregressive decoder; however the higher KL suggests that more information is encoded in the latents.
Our next experiment attempts to quantify how useful this information is.

\begin{table}
\caption{Binarized MNIST linear classification results}
\tlabel{mnist_classification_results}
\vskip 0.15in
\begin{center}
\begin{footnotesize}
\begin{sc}
\begin{tabular}{lr}
\toprule
Input & Accuracy (\%) \\
\midrule
PCA (16 components)        &  82.8\\
pixels     &  89.4\\
standard VAE codes &  95.4\\
Gated PixelVAE codes &  97.9\\
\textbf{ACN codes} &  $\textbf{98.5}$\\
\bottomrule
\end{tabular}
\end{sc}
\end{footnotesize}
\end{center}
\vskip -0.1in
\end{table}

\fig{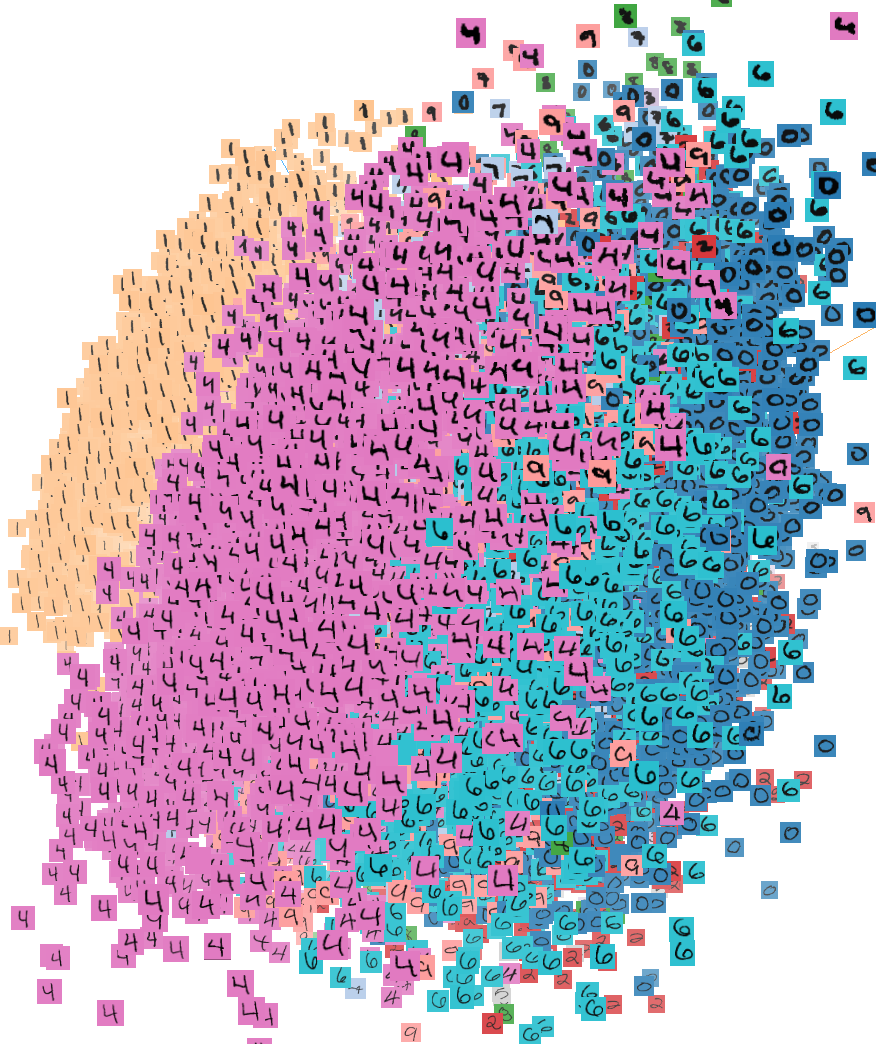}{mnist_pca}{1}{Visualisation of the first two principal components of ACN latent space for MNIST}{ Images are coloured according to class label~\cite{smilkov2016embedding}.}

\fig{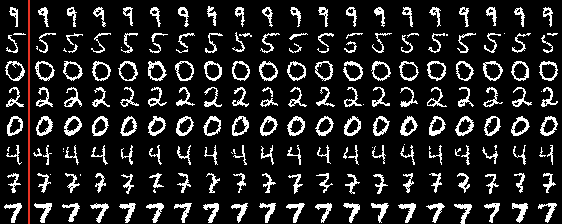}{bin_mnist_reconstruction}{1}{MNIST reconstructions}{ The codes for the test set images in the leftmost column were used to generate the samples in the remaining columns.}

\twofigstar{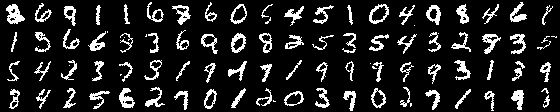}{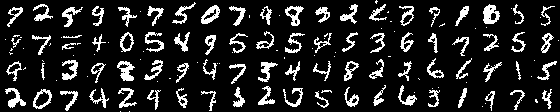}{bin_mnist_sample}{MNIST samples from ACN with unconditional prior (top) and Gated PixelCNN (bottom)}{}

\figstar{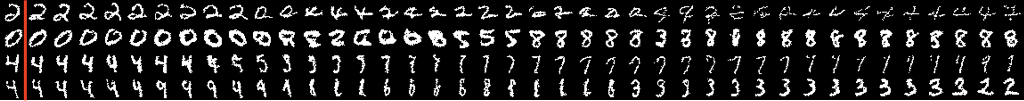}{bin_mnist_daydream}{1}{MNIST daydream samples}{ The leftmost column is from the test set. The remaining columns were generated by daydream sampling (\sref{sampling}).}

\tref{mnist_classification_results} shows the results of training a linear classifier to predict the training set labels with various inputs.
This gives us a measure of how the amount of easily accessible high-level information the inputs contain.
ACN codes are the most effective, but interestingly PixelVAE codes are a close second, in spite of having a KL cost of just over 1 nat per image.
VAE codes, with a KL of 26 nats per image, are considerably worse; we hypothesize that the use of a weaker decoder leads the VAE to include more low-level information in the codes, making them harder to classify.
In any case we can conclude that coding cost is not a reliable indicator of code utility.

The salience of the ACN codes is supported by the visualisation of the principal components of the codes shown in \fref{mnist_pca}: note the clustering of image classes (coloured differently to aid interpretation) and the gradation in writing style across the clusters (\eg strokes becoming thicker towards the top of the clusters, thinner towards the bottom). 
The reconstructions in \fref{bin_mnist_reconstruction} further stress the fidelity of digit class, stroke thickness, writing style and orientation within the codes, while the comparison between unconditional ACN samples and baseline samples from the Gated PixelCNN reveals a subtle improvement in sample quality.
\fref{bin_mnist_daydream} illustrates the dynamic modulation of daydream sampling as it moves through latent space: note the continual shift in rotation and stroke width, and the gradual morphing of one digit into another.

\subsection{CIFAR-10}

\fig{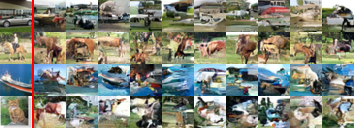}{cifar_reconstruction}{1}{CIFAR-10 reconstructions}{}

\twofigstar{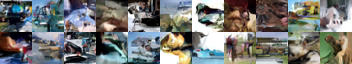}{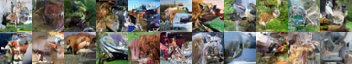}{cifar_sample}{CIFAR-10 samples from ACN with unconditional prior (top) and Gated PixelCNN (bottom)}{}

For the CIFAR-10 experiments the encoder was a convolutional network fashioned after a VGG-style classifier~\cite{simonyan2014very}, with 11 convolutional layers and 3x3 filters.
The decoder 
had 15 gated residual blocks, each using 128 filters of size 5x5; its output was a categorical distribution over subpixel intensities, with 256 bins for each colour channel.
Training batch size was 64.
The reconstructions in \fref{cifar_reconstruction} demonstrate some high level coherence, with object features such as parts of cars and horses occasionally visible, while \fref{cifar_sample} shows an improvement in sample coherence relative to the baseline.
We found that ACN codes for CIFAR-10 images were linearly classified with 55.3\% accuracy versus 38.4\% accuracy for pixels.
See Appendix A for more samples and results.

\subsection{ImageNet}

\figstar{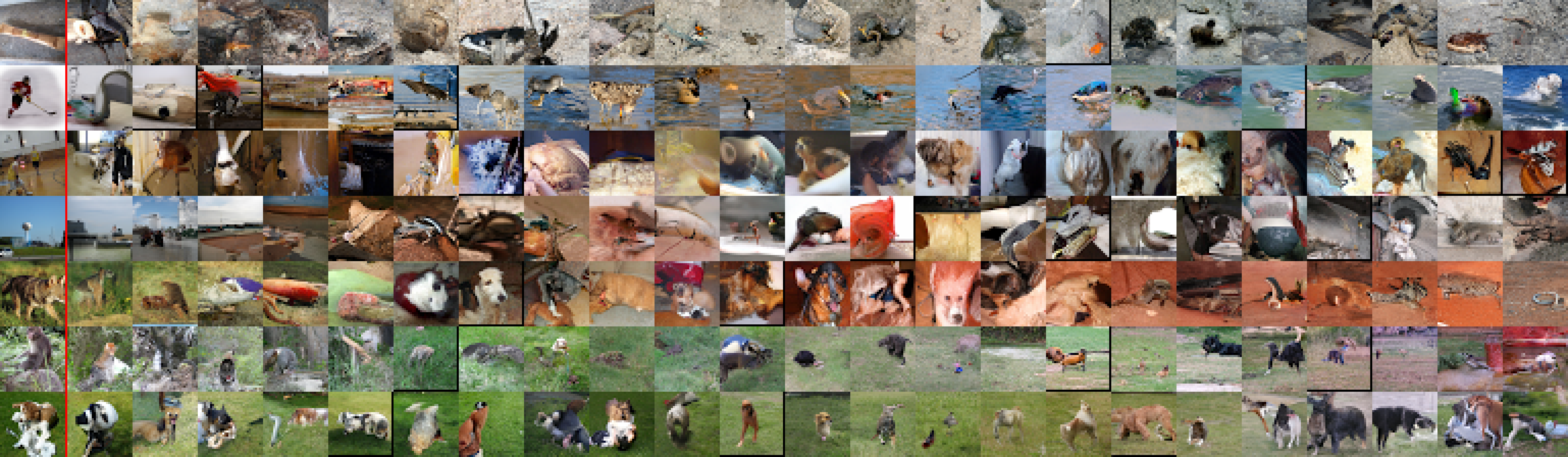}{imagenet_daydream}{1}{ImageNet daydream samples}{}

\fig{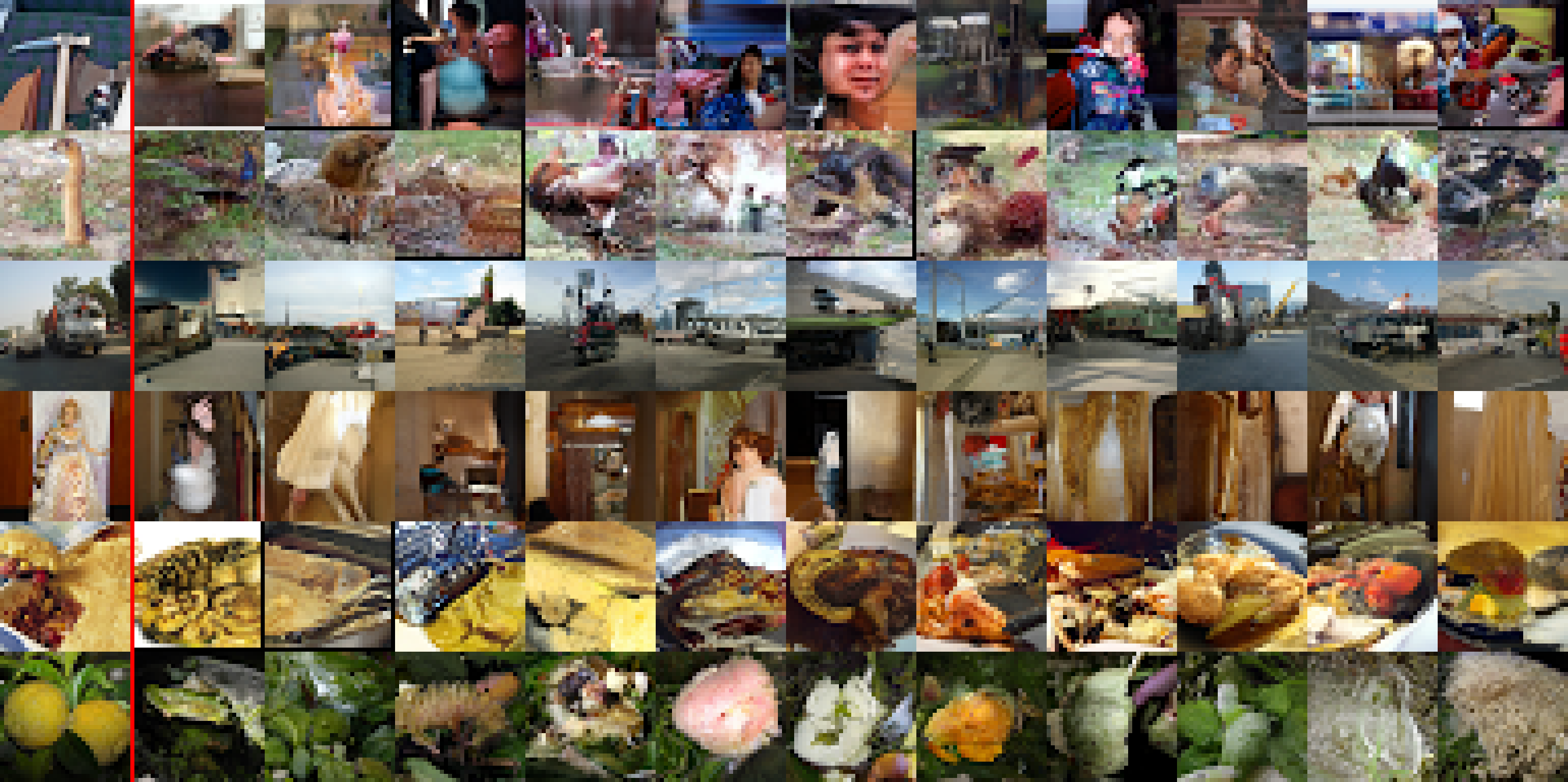}{imagenet_reconstruction}{1}{ImageNet reconstructions}{}

\twofigstar{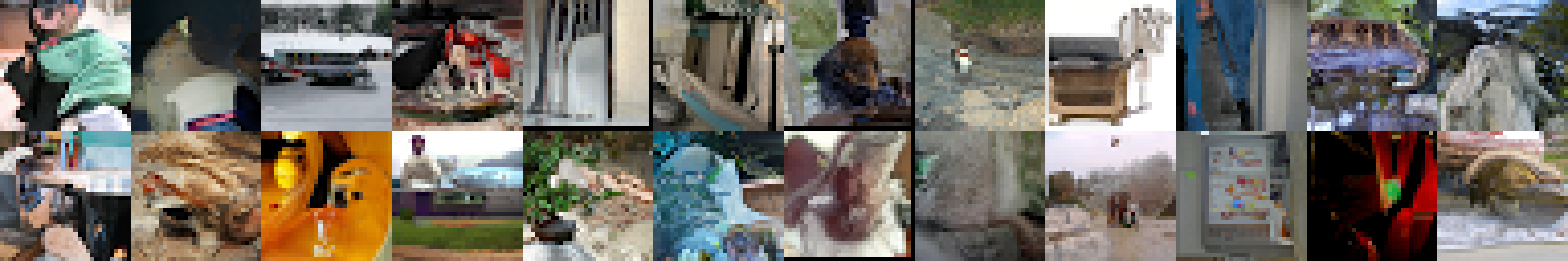}{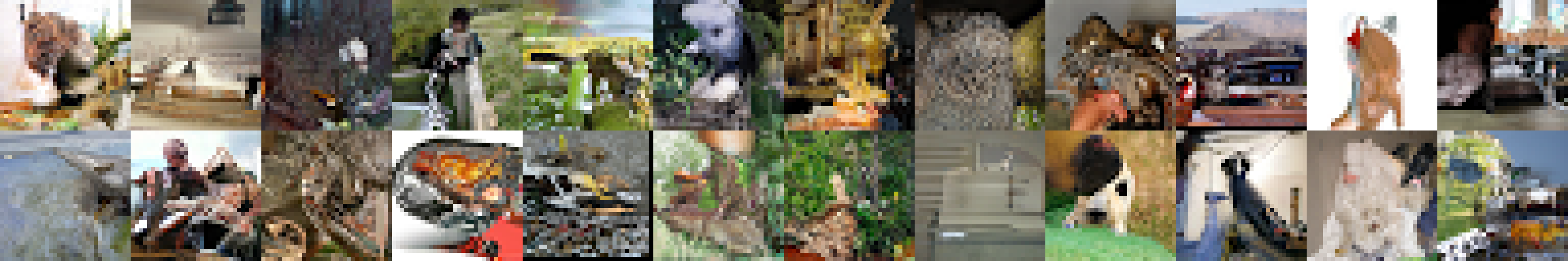}{imagenet_sample}{ImageNet samples from ACN with unconditional prior (top) and Gated PixelCNN (bottom)}{}

For these experiments the setup was the same as for CIFAR-10, except the decoder had 20 gated residual layers of 368 5x5 filters, and the batch size was 128.
We downsamples the images to 32x32 resolution to speed up training.
We found that ACN ImageNet codes can be linearly classified with 18.5\% top 1 accuracy and 40.5\% top 5 accuracy, compared to 3.0\% and 9.0\% respectively for pixels.
Better unsupervised classification scores have been recorded for ImageNet~\cite{doersch2015unsupervised,donahue2016adversarial,wang2015unsupervised}, but these were using higher resolution images.
The reconstructions in \fref{imagenet_reconstruction} suggest that ACN encodes information about image composition, colour, background and setting (natural, indoor, urban etc.), while \fref{imagenet_daydream} shows continuous transitions in background, foreground and colour during daydream sampling.
In this case the distinction between unconditional ACN samples and Gated PixelCNN samples was less clear (\fref{imagenet_sample}).
See Appendix B for more samples and results.
\subsection{CelebA}

\fig{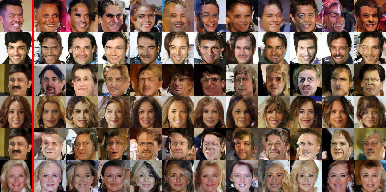}{celeba_reconstruction}{1}{CelebA reconstructions}{}

\twofigstar{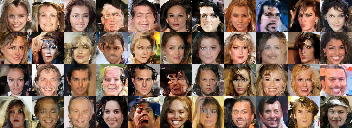}{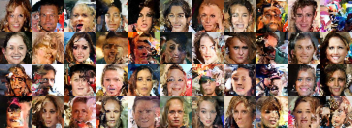}{celeba_samples}{CelebA samples from ACN with unconditional prior (top) and PixelCNN (bottom)}{}

We downsampled to CelebA images to 32x32 resolution and the same setup as for CIFAR-10.
\fref{celeba_reconstruction} demonstrates that high-level aspects of the original images, such as gender, pose, lighting, face shape and facial expression are well represented by the codes, but that the specific details are left to the decoder.
\fref{celeba_samples} demonstrates a slight advantage in sample quality over the baseline.

\section{Conclusion}
We have introduced Associative Compression Networks (ACNs), a new form of Variational Autoencoder in which associated codes are used to condition the latent prior.
Our experiments show that the latent representations learned by ACNs contain meaningful, high-level information that is not diminished by the use of autoregressive decoders.
As well as providing a clear conditioning signal for the samples, these representations can be used to cluster and linearly classify the data, suggesting that they will be useful for other cognitive tasks.
We have also seen that the joint latent and data space learned by the model can be naturally traversed by daydream sampling.
We hope this work will open the door to more holistic, dataset-wide approaches to generative modelling and representation learning.

\clearpage

\section*{Acknowledgements}
Many of our colleagues at DeepMind gave us valuable feedback on this work. We would particularly like to thank Andriy Mnih, Danilo Rezende, Igor Babuschkin,
John Jumper, Oriol Vinyals, Guillaume Desjardins, Lasse Espeholt, Chris Jones, Alex Pritzel, Irina Higgins, Loic Matthey, Siddhant Jayakumar and Koray Kavukcuoglu.

\bibliography{associative_compression_networks_arxiv}
\bibliographystyle{icml2018}

\clearpage

\appendix

\section{CIFAR-10}
\begin{table}[H]
\caption{CIFAR-10 test set compression results}
\tlabel{cifar_compression_results}
\vskip 0.15in
\begin{center}
\begin{footnotesize}
\begin{sc}
\begin{tabular}{lr}
\toprule
Model & Bits / dim \\
\midrule
DRAW~\cite{gregor2015draw} & 4.13\\
Conv DRAW~\cite{gregor2016towards} & 4.00\\
Pixel CNN~\cite{oord2016pixel}  & 3.14\\
Gated Pixel CNN~\cite{oord2016conditional}  & 3.03\\
Pixel RNN~\cite{oord2016pixel} & 3.00\\
PixelCNN++~\cite{salimans2017pixelcnn++} & 2.92\\
\textbf{PixelSNAIL}~\cite{chen2017pixelsnail} & \textbf{2.85}\\
\midrule
ACN (unordered)  & $3.07 $\\
\bottomrule
\end{tabular}
\end{sc}
\end{footnotesize}
\end{center}
\vskip -0.1in
\end{table}

\begin{table}[H]
\caption{CIFAR-10 test set ACN costs}
\tlabel{cifar_acn_costs}
\vskip 0.15in
\begin{center}
\begin{footnotesize}
\begin{sc}
\begin{tabular}{lr}
\toprule
Cost & Nats / image  \\
\midrule
KL ($K=1$)         & 5.4 \\
KL ($K=5$)         & 6.2 \\
KL (Tour)          & 6.3 \\
KL ($K=10$)        & 6.7 \\
KL (Unconditional) & 14.4 \\
\midrule
Reconstruction    & 6536.7 \\
\midrule
ACN (ordered)     & $\leq 6543.0$ \\
\bottomrule
\end{tabular}
\end{sc}
\end{footnotesize}
\end{center}
\vskip -0.1in
\end{table}

\twofigstarH{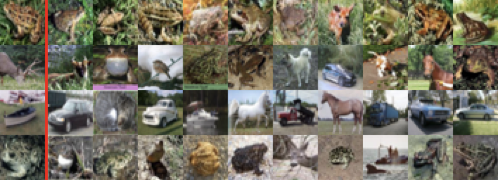}{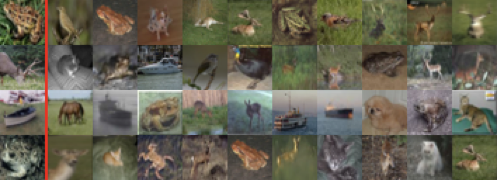}{cifar_neighbours}{CIFAR-10 nearest neighbours}{ The leftmost column is from the test set. The remaining columns show the nearest Euclidean neighbours in ACN code space (top) and pixel space (bottom) in order of increasing distance. While the codes often cluster according to high-level features such as object class and figure composition, clustering in pixel space tends to match on background colour, and disproportionately favours blurry images.}

\section{ImageNet}

\begin{table}[H]
\caption{ImageNet 32x32 test set compression results}
\tlabel{imagenet_compression_results}
\vskip 0.15in
\begin{center}
\begin{footnotesize}
\begin{sc}
\begin{tabular}{lr}
\toprule
Model & Bits / dim \\
\midrule
conv. DRAW~\cite{gregor2016towards} & 4.40\\
Pixel RNN~\cite{oord2016pixel} & 3.86\\
Gated Pixel CNN~\cite{oord2016conditional}  & 3.83\\
\textbf{PixelSNAIL}~\cite{chen2017pixelsnail} & \textbf{3.80}\\
\midrule
ACN (unordered)  & $3.82$\\
\bottomrule
\end{tabular}
\end{sc}
\end{footnotesize}
\end{center}
\vskip -0.1in
\end{table}

\begin{table}[H]
\caption{ImageNet test set ACN costs}
\tlabel{imagenet_acn_costs}
\vskip 0.15in
\begin{center}
\begin{footnotesize}
\begin{sc}
\begin{tabular}{lr}
\toprule
Cost & Nats / image  \\
\midrule
KL ($K=1$)         & 2.9 \\
KL ($K=5$)         & 8.7 \\
KL ($K=10$)        & 10.3 \\
KL (Greedy Tour)   & 10.6 \\
KL (Unconditional) & 18.2 \\
\midrule
Reconstruction    & 8112.8 \\
\midrule
ACN (ordered)     & $\leq$ 8123.4 \\
\bottomrule
\end{tabular}
\end{sc}
\end{footnotesize}
\end{center}
\vskip -0.1in
\end{table}

\figstar{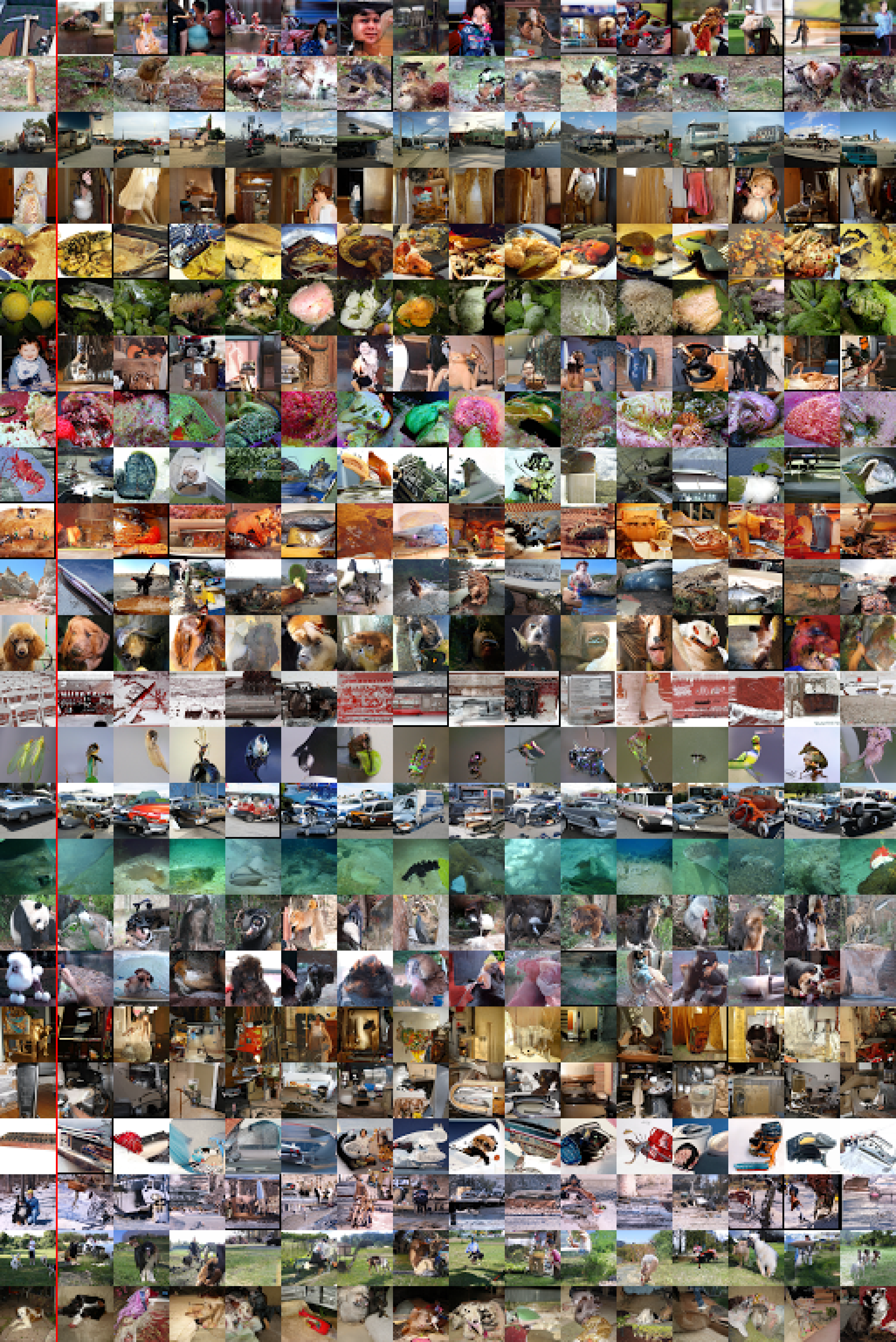}{imagenet_reconstruction_full_page}{0.85}{ImageNet reconstructions}{}

\figstar{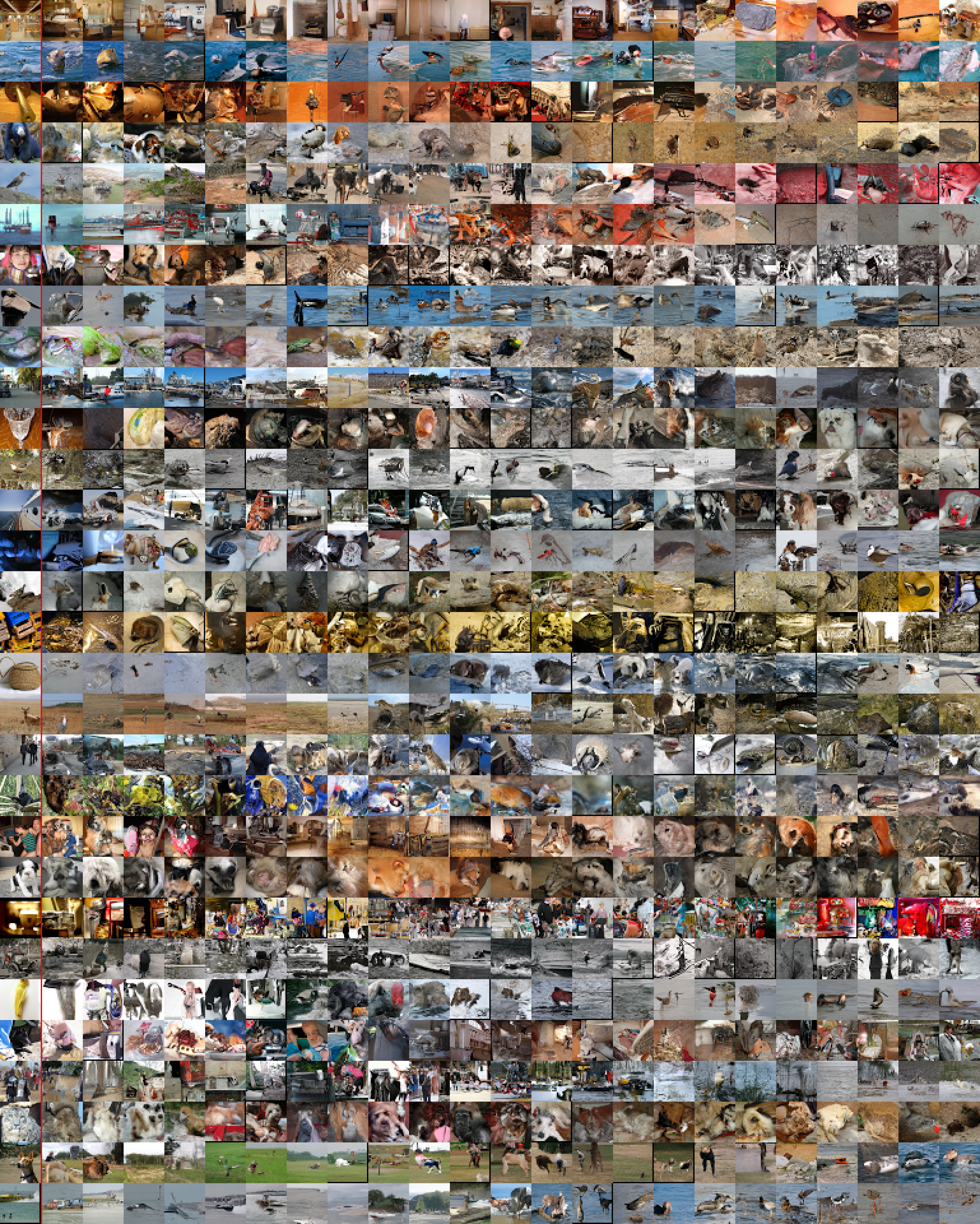}{imagenet_daydream_full_page}{1}{ImageNet daydream samples}{}

\end{document}